\documentclass[letterpaper]{article} 
\usepackage{aaai25}  
\usepackage{times}  
\usepackage{helvet}  
\usepackage{courier}  
\usepackage[hyphens]{url}  
\usepackage{graphicx} 
\usepackage{natbib}  
\usepackage{caption} 
\usepackage{algorithm}
\usepackage{algorithmic}

\usepackage{newfloat}
\usepackage{listings}
\DeclareCaptionStyle{ruled}{labelfont=normalfont,labelsep=colon,strut=off} 
\floatstyle{ruled}
\newfloat{listing}{tb}{lst}{}
\floatname{listing}{Listing}
\usepackage{breqn}
\usepackage{verbatim}
\usepackage{fvextra}
\usepackage{booktabs}
\usepackage{makecell}
\usepackage{amsmath}

\usepackage{color}
\usepackage[T1]{fontenc}
\usepackage{textcomp}
\usepackage{amsfonts}
\usepackage{amsthm}

\title{Instruction-Augmented Long-Horizon Planning: Embedding Grounding Mechanisms in Embodied Mobile Manipulation}
\author {
    Fangyuan~Wang\textsuperscript{\rm 1,\rm 2},
    Shipeng~Lyu\textsuperscript{\rm 1, \rm 2},
    Peng~Zhou\textsuperscript{\rm 3},
    Anqing~Duan\textsuperscript{\rm 4}, \\
    Guodong~Guo\textsuperscript{\rm 2}\footnotemark[1],
    David~Navarro-Alarcon\textsuperscript{\rm 1}\thanks{Corresponding authors}
}
\affiliations {
    \textsuperscript{\rm 1} The Hong Kong Polytechnic University
    \textsuperscript{\rm 2} Ningbo Institute of Digital Twin, Eastern Institute of Technology\\
    \textsuperscript{\rm 3} Great Bay University
    \textsuperscript{\rm 4} Mohamed Bin Zayed University of Artificial Intelligence\\
    gdguo@eitech.edu.cn, 
    dnavar@polyu.edu.hk
}

\usepackage{bibentry}

\begin{document}

\maketitle

\begin{abstract}
Enabling humanoid robots to perform long-horizon mobile manipulation planning in real-world environments based on embodied perception and comprehension abilities has been a longstanding challenge. 
With the recent rise of large language models (LLMs), there has been a notable increase in the development of LLM-based planners. These approaches either utilize human-provided textual representations of the real world or heavily depend on prompt engineering to extract such representations, lacking the capability to quantitatively understand the environment, such as determining the feasibility of manipulating objects.
To address these limitations, we present the Instruction-Augmented Long-Horizon Planning (IALP) system, a novel framework that employs LLMs to generate feasible and optimal actions based on real-time sensor feedback, including grounded knowledge of the environment, in a closed-loop interaction. Distinct from prior works, our approach augments user instructions into PDDL problems by leveraging both the abstract reasoning capabilities of LLMs and grounding mechanisms.
By conducting various real-world long-horizon tasks, each consisting of seven distinct manipulatory skills, our results demonstrate that the IALP system can efficiently solve these tasks with an average success rate exceeding 80\%.
Our proposed method can operate as a high-level planner, equipping robots with substantial autonomy in unstructured environments through the utilization of multi-modal sensor inputs. 
\end{abstract}

\begin{links}
\link{Website}{https://nicehiro.github.io/IALP}
\end{links}

\section{Introduction}
Embodied decision-making has been regarded as a critical role for humanoid robots tasked with executing long-horizon mobile manipulation tasks, which require prolonged navigation of the environment and interaction with objects.
The recent advancement of large language models (LLMs) has significantly enhanced their ability to perceive, comprehend, and plan actions with their surroundings, thus promoting the successful completion of long-horizon mobile manipulation tasks.
However, most of the existing research lacks the grounding knowledge of the real-world environment, such as spatial positioning and feasible grasping poses. 
They either rely on simulated settings or predefined contextual descriptions, with few investigations into real-world understanding and representation through the use of LLMs. 
This deficiency in grounded understanding impedes robots from properly interacting with complex objects/structures, thus, impairing their application in unknown environments.
Given these challenges, an imperative question arises: Can we effectively extract and utilize environmental representations, both abstract and grounded, to empower humanoid robots to plan long-horizon tasks in real-world environments?

\begin{figure}[t]
  \centering
  \includegraphics[width=\columnwidth]{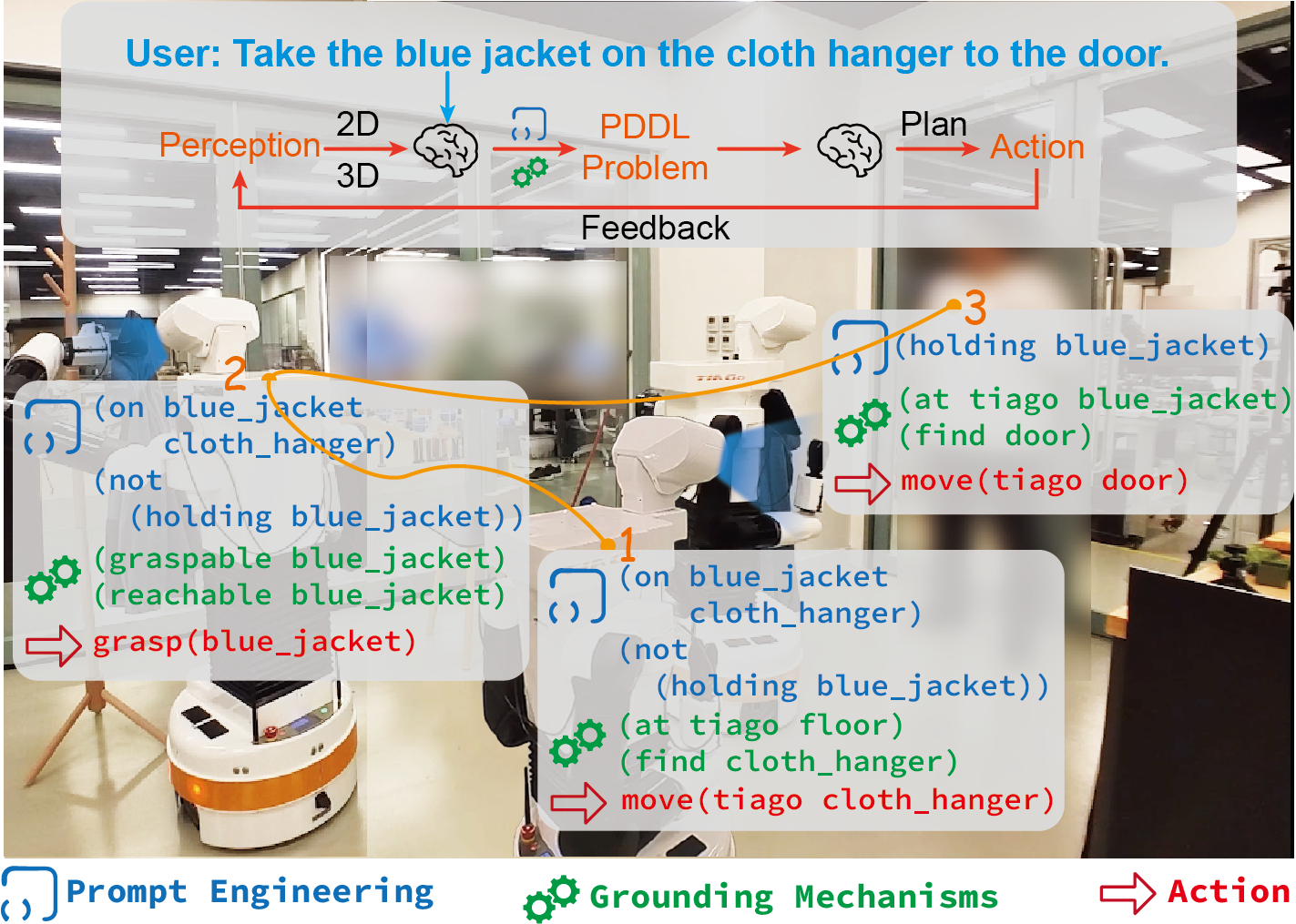}
  \caption{The IALP system leverages the reasoning capabilities of LLMs and grounding mechanisms to enrich the task representation, enabling plan feasible and optimal actions in the real world.}
  \label{fig:intro}
\end{figure}

An approach to providing robots with embodied intelligence involves utilizing off-the-shelf foundation models such as GPT-4 \cite{openai2024gpt4technicalreport}, CLIP \cite{radford2021learning}, and SAM \cite{kirillov2023segment}, thus enabling them to do long-horizon planning. 
Classical methods \cite{wake_chatgpt_2023, lin_text2motion_2023, huang2023instruct2act, ahn2024autort} typically operate within simulation environments or assume the availability of textual descriptions of the environment, heavily relying on the reasoning capabilities of visual language models (VLMs) through prompt engineering. However, these methods face significant limitations in real-world environments where extracting key features, such as the precise position of objects, from the real world is challenging.

Furthermore, mobile manipulation in robots has gained increasing popularity. Previous studies have employed both models trained via transformer-based structures \cite{brohan2022rt, brohan2023rt} and training-free approaches \cite{liu2024ok, wang2024hypermotion, ahn2024autort}. Nonetheless, these approaches either lack long-horizon planning capabilities \cite{brohan2022rt, brohan2023rt, liu2024ok} and/or interactive planning abilities that enable the robot to adjust its behavior based on feedback \cite{liu2024ok, wang2024hypermotion, ahn2024autort}.
We anticipate leveraging the implicit world knowledge of LLMs and grounding mechanisms, such as voxel map building and grasping pose generation, to decode task-relevant information from the environment. This information can be transformed into structured descriptions that are recognized by the LLM planner, enabling the execution of long-horizon tasks in a closed-loop manner.

To this end, we introduce the \textbf{I}nstruction \textbf{A}ugmented \textbf{L}ong-Horizon \textbf{P}lanning (\textbf{IALP}) system: a flexible and extendable framework to solve long-horizon tasks in real-world environments. The IALP system (i) augments the user instruction with the task-related knowledge extracted from promptable and grounding mechanisms-based predicates using Planning Domain Description Language (PDDL) \cite{fox2003pddl2, garrett2020pddlstream}, and (ii) generates feasible and optimal actions to complete long-horizon tasks in a \emph{close-loop} manner (see Figure \ref{fig:intro}). 
Specifically, our new method first augments the user's instruction by generating a PDDL problem based on the user's input and the robot's perception, which includes both 2D and 3D data. 
Then, the system employs the LLM planner to interactively generate feasible and optimal actions, encompassing both navigation and manipulation, to accomplish the entire task. 
The system's extendability allows for the design and inclusion of additional predicates, both promptable and grounding mechanisms, to meet various needs. The contributions are summarized as follows:

\begin{itemize}
  \item We introduce IALP, an adaptive framework that utilizes both abstract reasoning abilities (from LLMs) and grounding mechanisms to extract task-relevant representations from the environment.
  \item We propose a method that enables robots to generate feasible and optimal actions based on real-time feedback in a closed-loop interaction with the environment.
  \item We deploy the entire system in real-world scenarios across several long-horizon tasks, comprising skills such as moving, picking, placing, and more. The results demonstrate that our algorithm achieves over 80\% planning accuracy.
\end{itemize}

\section{Related Works}
\subsection{Pretrained Foundation Models for Task Planning}
Preliminary studies \cite{kambhampati2023role, valmeekam2023planning, guan2023leveraging, ao2024llm} have demonstrated that off-the-shelf large language models (LLMs) have the capability to solve long-horizon tasks as effective high-level planners. Traditional planning systems (e.g., those based on PDDL) employ efficient search algorithms to identify correct or even optimal plans, but are constrained by predefined structured task descriptions. Recent studies \cite{liu2023llm+, silver2022pddl} have leveraged the capabilities of LLMs and PDDL planners, where LLMs are employed to generate structured task descriptions, and PDDL planners subsequently used to find optimal solutions.

With the integration of multimodal inputs, such as visual and auditory data, several studies \cite{geng2023sage,zhang2024dkprompt,huang2023instruct2act} have heavily relied on the reasoning and in-context learning capabilities of visual language models (VLMs) to extract spatial relations and object attributes for planning purposes. However, we argue that while VLMs excel at reasoning about the spatial relations in 2D images, they are limited in their ability to extract precise spatial information, such as an object's position and pose within the environment.
Rather than relying on predefined planning priors, we leverage not only VLMs' implicit world knowledge but also grounding mechanisms to extract representations that encode task-relevant information, such as checking if the generated grasping pose is feasible to reach.

\subsection{Mobile Manipulation in Open Worlds}
Mobile manipulation and navigation in open-world environments necessitate open-vocabulary capabilities and strong adaptive skills in robots. Despite recent advances in pretraining foundation models, which have spurred increased research interest, this remains an unresolved challenge \cite{yenamandra2023homerobot}. To address this issue, \cite{brohan2022rt, brohan2023rt} propose end-to-end pipelines by training transformer-based vision-language models. On the other hand, \cite{liu2024ok, wang2024hypermotion, ahn2024autort} introduce training-free approaches utilizing pretrained foundation models.
OK-Robot \cite{liu2024ok} can perform pick-and-place tasks in open-world settings without any training by leveraging state-of-the-art vision-language models (VLMs) such as CLIP \cite{radford2021learning} and OWL-ViT \cite{minderer2022simple} for object detection, navigation and manipulation. However, it lacks the ability to reason about long-horizon tasks and is limited to pick-and-place operations. In contrast, \cite{wang2024hypermotion, ahn2024autort} use distilled spatial geometry and 2D observations with VLMs to guide the morphology selection of robots. Their proposed planning frameworks are one-shot, meaning they plan the entire task at the beginning without receiving feedback and use LLMs only for morphology selection rather than for generating primitive actions.
Prior studies either lack long-horizon planning capabilities \cite{brohan2022rt, brohan2023rt, liu2024ok} or interactive planning abilities \cite{wang2024hypermotion, ahn2024autort}.

Our work constructs the PDDL problem description based on the observation of the current state using embodied perception. These textual descriptions are then fed into LLMs. After executing the generated action, the current observation changes, prompting the reconstruction of the PDDL problem. This process is iteratively executed, enabling our method to achieve both long-horizon planning and interactive planning capabilities.

\section{Problem Formulation}
We aim to address long-horizon mobile manipulation tasks that necessitate both symbolic and geometric reasoning from a natural language instruction $i$ and the initial state of the environment. We assume an open-world setting, wherein the robot operates without prior knowledge of task-relevant objects or other ground truth information. Instead, it relies solely on data acquired through its embodied perception system and a prebuilt voxel map for navigation.

We utilize an LLM planner with a library of skills $L^{\psi} = \{\psi_1, \psi_2, \cdots, \psi_n\}$. Each skill $\psi$ is associated with a single-timestep Markov Decision Process (MDP) \cite{sutton2018reinforcement, lin_text2motion_2023}, which is denoted as $(\mathcal{S}, \mathcal{A}, R, T, \rho)$. $\mathcal{S}$ is the state space, $\mathcal{A}$ is the action space, $T: \mathcal{S} \times \mathcal{A} \rightarrow \mathcal{S}$ is the transition function, $R$ is the reward function, and $\rho$ is the distribution of the initial state.
Each action $a$ is a skill $\psi \in L^{\psi}$, which is a parameterized primitive $\phi(\alpha)$.
When an action is executed, a parameter $\alpha$ is sampled from the LLM planner and fed to its primitive $\phi(\alpha)$, which executes a series of motor commands on the robot. If the action succeeds, it receives a binary reward of $r$.

\subsection{The Planning Objective}
The objective is to find a sequence of actions $\{a_1, \cdots, a_H\}$, denoted as $a_{1:H}$, that can achieve the given instruction $i$.
We use an interactive planning method that considers action feedback and proposes several candidates at each time step.
Thus, the objective at time step $t \in \{1:H\}$ can be expressed as the joint probability of skill sequence $a_{t:H}$ and binary rewards $r_{t:H}$ given the instruction $i$ and the current state $s_t$:

\begin{equation}
\begin{aligned}
\label{eq:pf-1}
  &p(a_{t:H}, r_{t:H}|i, s_t) = p(a_{t:H} \mid i, s_t) p(r_{t:H} \mid i, s_t, a_{t:H}) \\
                           &= \prod_{x=t}^H p(a_x \mid i, s_{t:x}, a_{t:x-1}) \prod_{x=t}^{H} p(r_x \mid s_{t:x}, a_{t:x}),
\end{aligned}
\end{equation}
where the former term represents the probability that the action sequence $a_{t:H}$ will satisfy the instruction $i$ from a symbolic perspective.
We define the action sequence \textbf{optimality} score $S_{op} = \prod_{x=t}^H p(a_x \mid i, s_{t:x}, a_{t:x-1})$, where the probability of the next skill $a_x$ is considered in terms of the current state $s_t$, predicted $s_{t+1:x}$ and planned $a_{t:x-1}$ and is thus independent of prior rewards $r_{1:x-1}$. It captures the utility of the action sequence $a_{t:H}$ with respect to satisfying the instruction $i$ on current state $s_t$.

The later term of Equation \ref{eq:pf-1} represents the probability that the action sequence $a_{t:H}$ achieve rewards $r_{t:H}$ when executed from the state $s_t$, which is conditionally independent of the instruction $i$.
We consider the success probability of the action sequence $\prod_{x=t}^H p(r_x \mid s_{t:x}, a_{t:x})$ as \textbf{feasibility} score $S_{fb}$.
The sequence of actions $\{a_t, \cdots, a_H\}$ is feasible if the robot can approach each action successfully, i.e., the reward sequence $r_{t:H}$ are all equal to 1.
If even one skill fails, then the entire action sequence fails.

\section{Methodology}
\begin{figure*}
  \centering
  \includegraphics[width=\textwidth]{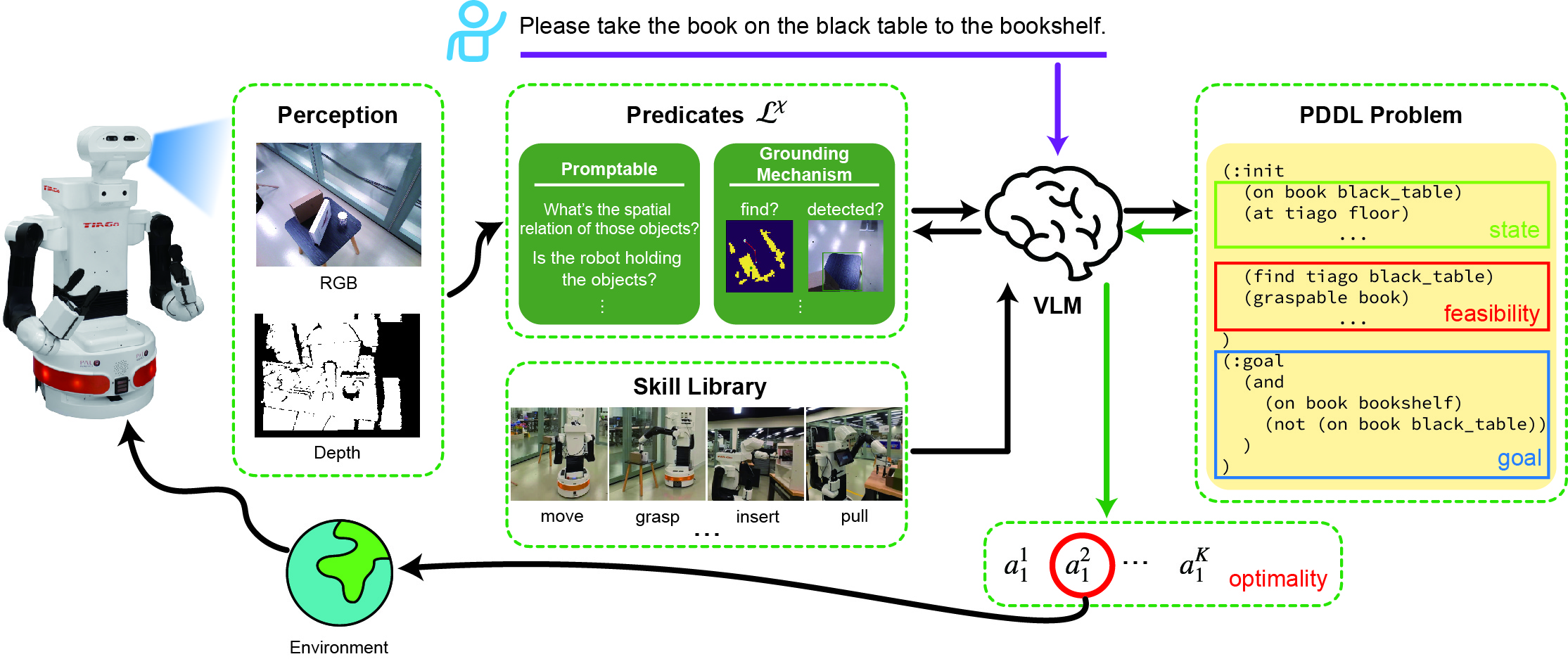}
  \caption{The proposed IALP system is designed to complete long-horizon mobile manipulation tasks in real world environment. Firstly, it constructs a PDDL problem for the task at current state by utilizing promptable and ground truth predicates derived from embodied perception and the skill library. Then, the robot executes the actions generated and selected by the LLM planner based on the constructed PDDL problem. The system operates in a closed-loop manner until the task is completed.}
  \label{fig:overview}
\end{figure*}

As illustrated in Figure. \ref{fig:overview}, we propose the Instruction-Augmented Long-Horizon Planning (IALP) system to infer actions for solving long-horizon tasks in a close-loop manner.
Firstly, the system augments the user instruction into a PDDL problem through promptable and grounding mechanism-based predicates. These predicates utilize 2D and depth information obtained from embodied sensors, along with a primitive skill library $L^{\psi}$, to represent the current state. Based on augmented instruction, we utilize the LLM planner to generate $K$ action candidates, from which the optimal action is selected based on the highest token probability.
The system's extensibility and flexibility are demonstrated by the capability to incorporate additional promptable and grounding mechanism-based predicates as required by the task.

\subsection{Instruction Augmentation}
To provide efficient and sufficient task-related knowledge for the planner, we leverage off-the-shelf visual foundation models, specifically the SAM and CLIP, among others, accessed via designated APIs. This approach facilitates the construction of the PDDL problem by augmenting the textual human instruction $i$ with embodied sensors.
The LLM-based planner requires a domain description $D$ that defines a set of predicates (e.g., \verb|reachable|) and actions (e.g., \verb|move|), as well as a problem description $P$ for each task. The problem description defines the task by specifying the \verb|init| state and \verb|goal| state using a list of predicates defined in the domain file.

\begin{table}[t]
  \centering
  \begin{tabular}{rc}
    \toprule
    Promptable & on, in, holding, opened \\
    \midrule
    Grounding Mechanism & \makecell{at, find, graspable, placeable, \\ detected, reachable} \\
    \bottomrule
  \end{tabular}
  \caption{Promptable and grounding mechanism-based predicates.}
  \label{tab:predicates}
\end{table}

\noindent\textbf{Predicates.} We consider utilizing a predicate library $\mathcal{L}^{\chi} = \{\chi^1, \cdots, \chi^M\}$, as shown in Table \ref{tab:predicates}, to describe the current observation and feasibility feedback.
Each predicate is a binary-valued function over objects. For example, \verb|find| represents a binary-valued function, with \verb|book| serving as the object argument for predicate \verb|(find book)|.
This library consists of four promptable predicates that can be addressed through prompt engineering based on the reasoning ability of state-of-the-art LLMs, such as \verb|holding| and \verb|on|, and six predicates determined by grounding mechanisms, such as \verb|find| and \verb|reachable|. 
For promptable predicates, LLMs will generate ``True'' or ``False'' based on textual or multi-modal inputs prompted by designed prompts. For example, if the robot successfully takes the book from the table, LLMs might respond ``False'' to \verb|(on book table)| and ``True'' to \verb|(holding book)|. 
Grounding mechanism-based predicates are primarily employed for generating feasibility feedback, as detailed in Section Feasibility.


\setlength{\tabcolsep}{1mm}
\SaveVerb{obj}|obj|
\SaveVerb{pla}|pla|
\begin{table}[ht]
  \small
  \centering
  \begin{tabular}{lll}
    \toprule
    Actions & Preconditions $\psi^p$ & Effects $\psi^e$ \\
    \midrule
    \verb|move| & \verb|(find obj)| & \verb|(at obj)| \\
    \hline
    \verb|scan| & \makecell[l]{\Verb|(at obj)| \\ \Verb|(not| \\ \quad\Verb|(detected obj))|} & \verb|(detected obj)| \\
    \hline
    \verb|grasp| & \makecell[l]{\Verb|(at obj)| \\ \Verb|(detected obj)| \\ \Verb|(reachable obj)| \\ \Verb|(graspable obj)|} & \verb|(holding obj)| \\
    \hline
    \verb|place| & \makecell[l]{\Verb|(at pla)| \\ \Verb|(holding obj)| \\ \Verb|(peaceable pla)| \\ \Verb|(reachable pla)|} & \makecell[l]{\Verb|(not| \\ \quad\Verb| (holding obj))| \\ \Verb|(on obj pla)|} \\
    \hline
    \verb|lift| & \verb|(holding obj)| & \verb|(holding obj)| \\
    \hline
    \verb|pull| & \verb|(holding obj)| & \verb|(opened obj)| \\
    \hline
    \verb|insert| & \makecell[l]{ \Verb|(at pla)| \\ \Verb|(holding obj)| \\ \Verb|(peaceable pla)| \\ \Verb|(reachable pla)|} & \makecell[l]{\Verb|(not| \\ \quad\Verb| (holding obj))| \\ \Verb|(on obj pla)|} \\
    \bottomrule
  \end{tabular}
  \caption{Preconditions and effects of actions in PDDL. \protect\UseVerb{obj} and \protect\UseVerb{pla} refer to the name of the object and peaceable space involved in the action.}
  \label{tab:actions}
\end{table}

\noindent\textbf{Actions.} The robot is equipped with a primitive action library $\mathcal{L}^{\psi}$, which is inherently imperfect and frequently cause unforeseen situations. We list seven actions in Table \ref{tab:actions} each associated with a preconditions set $\psi^p$ and effects $\psi^e$ of executing the action. 
An action is feasible to execute if and only if all the predicates in its precondition set are evaluated as ``True''.
A reward of 1 is received if the robot successfully executes with the effects representing the expected state changes.
For example, \verb|move(obj)| denotes the moving action with a primitive function $\phi$: \verb|move| and its parameter $\alpha$: \verb|obj|. This action can be executed only if there's at least one path from the current position to the \verb|obj|.

\noindent\textbf{Current state.} Unlike the definition in the original PDDL paper, which uses \verb|:init| as the initial state of the task, we utilize it to represent the current state $s_t$ at every timestep $t$. We initially employ LLMs to identify task-specific objects relevant to the current task. For instance, given the instruction ``Take the blue jacket on the cloth hanger to the door'', the objects ``blue jacket'', ``cloth hanger'' and ``door'' are extracted. Subsequently, utilizing these identified objects and the provided instruction, we apply the list of promptable predicates (Table \ref{tab:predicates}) to derive knowledge about the current state. This includes determining the spatial relationships among the objects and the status (e.g., whether an object is being held or if a drawer is open) of the manipulated object. 

\noindent\textbf{Goal state.} Given an instruction $i$, a set of objects $o$ to be used in the current task is first generated by the LLM. Based on predicates library $\mathcal{L}^{\chi}$, we use the LLM to predict a symbolic goal $g$, which is a set of predicates grounded over objects $o$ in the scene, to satisfy the instruction. 

\subsection{Feasibility}
We employ an interactive planning approach, which involves replanning at each time step to allow the planner to adapt even if previous actions failed. Thus, here we consider the feasibility score of planning single time step, i.e., $S_{fb} = p(r_t|s_t, a_t)$.
We aim to maximize the probability $p(1|s_t, a_t)$ of achieving a reward of 1 at state $s_t$ by performing action $a_t$, as determined by the LLM planner. The entire action sequence $a_{1:H}$ fails, denoted by $S_{fb} = 0$, if at least one action fails, i.e., $r_t = 0, \exists t \in \{1:H\}$.
Two primary factors influence the probability $p(1|s_t, a_t)$. First, the feasibility of executing the action $a_t$ at state $s_t$, such as whether the object to be manipulated can be grasped. Second, the likelihood of the action $a_t$ succeeding when it is feasible. For instance, a robot cannot move toward a blue jacket if it cannot identify a feasible path. When there's a feasible path, navigation errors can still cause the action to fail.

To address the first factor, we aim to enhance the probability of action feasibility by providing more detailed feasibility feedback. 
We introduce six feasibility predicates, comprising two navigation predicates and four manipulation predicates, to maximize the feasibility score $S_{fb}$ thereby increasing the likelihood that the generated actions can be successfully executed.

\noindent\textbf{Navigation feasibility.} Upon receiving an action to move toward an object, the robot must ensure the existence of at least one viable path from its current position to the target object.
We utilize the predicate \verb|find| to represent this determination and \verb|at| to record the robot's current position.

To achieve this, we constructed a voxel map to provide a static representation of the current environment, employing the natural language mapping approach proposed by \cite{chen2023open, liu2024ok}. This method maps visual-language embeddings of objects to their corresponding positions.
We then use the SAM to generate masks for all potential objects (extracted from the current state) within the environment. From those masks, we extract corresponding crops $c$, including 2D image crops $c_{RGB}$ and depth crops $c_D$.
Object embeddings $e_{o}$ are obtained via CLIP to encode image crops $c_{RGB}$, while the corresponding position $(x_o, y_o, z_o)$ are derived from depth crops $c_D$.
For a given object language description $l_o$, we first convert it to a semantic vector $e_o^{\prime}$ using the CLIP text encoder.
The spatial position of an object language description $l_o$ can be extracted by sampling from the voxel map, where the dot product between the image-encoded vector $e_o$ and text-encoded vector $e_o^{\prime}$ is maximized.
The $A^{*}$ algorithm is employed to identify a path from the current position to the target position.
For \verb|at| predicate, we assume the robot will reach the target position after executing the action to move to the object.

\noindent\textbf{Manipulation feasibility.} We employ predicates \verb|detected|, \verb|graspable|, \verb|placeable| and \verb|reachable| to evaluate whether an object can be manipulated.

When the robot arrives at the object, we first ensure that the object is within the camera's field of view, denoted by predicate \verb|detected|. To achieve this, we utilize the LangSAM \cite{medeiros2023lang} model, which extracts the desired object's mask based on the object language description $l_o$ and the image captured by the robot.
For determining the predicates \verb|graspable| and \verb|placeable|, we employ the grasping and placing heuristics proposed by \cite{liu2024ok} which utilize foundational VLMs to verify and generate graspable poses and placeable areas. A pre-trained GraspNet \cite{fang2023anygrasp} is used to generate several potential grasps for the robot, with LangSAM masks filtering the object-related grasps. The optimal placeable location for various surfaces is determined by first segmenting and aligning the point cloud, then calculating median coordinates and identifying the maximum height of the placeable object.
For predicate \verb|reachable|, we query these grasp poses in the pre-computed reachability map, where the reachability of any SE(3) end-effector pose of a robot is determined \cite{jauhri2024active, zacharias2007capturing, vahrenkamp2013robot}. We exclude any grasps that cannot be reached in the current state by computing a grasp reachability index for each candidate grasp. This map is created by executing forward kinematics for numerous joint configurations and recording the visitations and manipulability of the 6D voxels visited by the end-effector.

\subsection{Optimality}
We aim to maximize the optimality score $S_{op}$ of the action sequence $a_{t:H}$ at time step $t$. 
Given our interactive planning method, which involves planning actions at every time step, the objective of the optimality score for planning at a single time step in state $s_t$ is to maximize the probability $p(a_t | i, s_t)$ of next action $a_t$. This probability is independent of prior rewards $r_{1:t-1}$ and actions $a_{1:t-1}$.
Given the PDDL problem $P$ of a specific task, domain $D$, and user instruction $i$, we first query the LLM planner to generate $K$ candidates actions $\{a_t^1, \cdots, a_t^K\}$ at state $s_t$. We then compute the optimality scores $S_{op}$ by summing the token log probabilities of each action's language description. These scores represent the likelihood that $a_t^k$ is the optimal skill to execute from a language modeling perspective to satisfy instruction $i$.
The action with the highest token probability is then selected as the optimal action, i.e.,

\begin{equation}
a_t^* = \arg\max p_{t}(a_t^k | i, D, P), k \in \{1,\cdots, K\}.
\end{equation}

where $p_{t}$ is token probability of action.

\section{Experiments}
\subsection{Experimental Setup}

\begin{figure}[t]
    \centering
    \includegraphics[width=0.8\columnwidth]{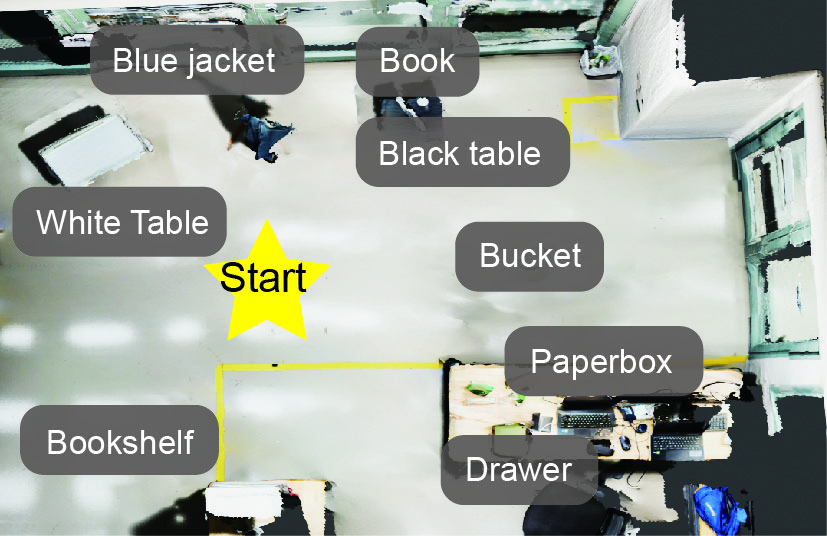}
    \caption{The room environment we used in our tasks. The position of the yellow star is the initial state of the robot.}
    \label{fig:room}
\end{figure}


\begin{table}[tbp]
  \centering
  \begin{tabular}{cc}
    \toprule
    Task & Instruction \\
    \midrule
    Place box & \makecell[l]{Pick the paper box on the wooden table \\ and place it on the black table.} \\
    \hline
    Take jacket & \makecell[l]{Take the blue jacket on the cloth hanger \\ to the door.} \\
    \hline
    Take pillbox & \makecell[l]{Take the pill box in the drawer to me.} \\
    \hline
    Insert book & \makecell[l]{Take the book from the black table and \\ put it on the bookshelf.} \\
    \hline
    Lift bucket & \makecell[l]{Lift the blue bucket to the white table.} \\
    \bottomrule
  \end{tabular}
  \caption{The five long-horizon tasks and instructions employed in our experiments.}
  \label{tab:scene}
\end{table}


\begin{figure*}[t]
  \centering
  \includegraphics[width=\textwidth]{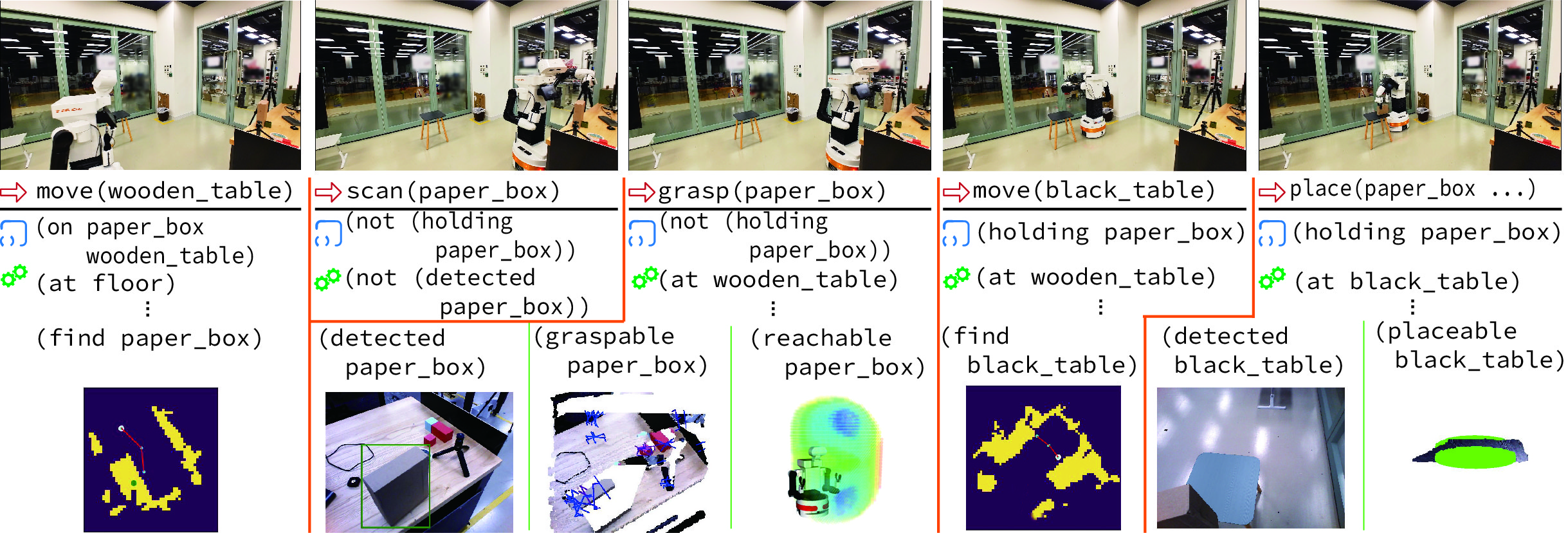}
  \caption{The states, feasibility feedback, and actions during the execution of long-horizon mobile manipulation tasks.}
  \label{fig:plans-box}
\end{figure*}

\begin{figure*}[ht]
  \centering
  \includegraphics[width=\textwidth]{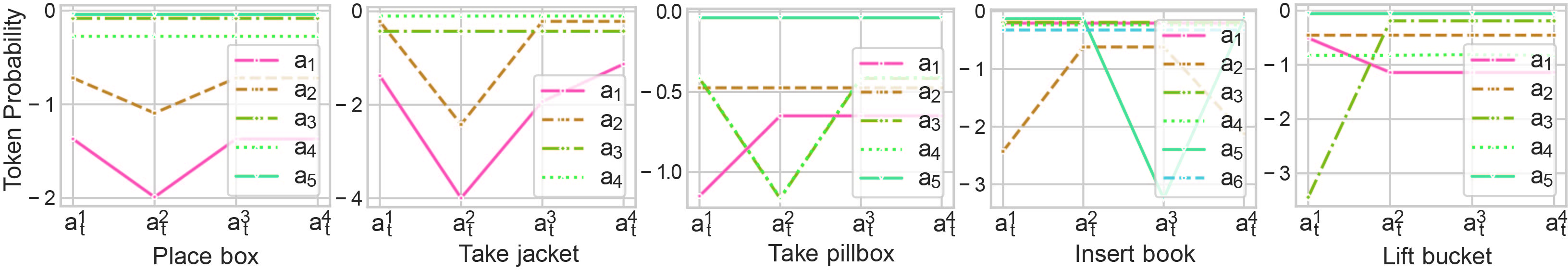}
  \caption{The token probability of action candidates generated by the LLM planner for five long-horizon tasks.}
  \label{fig:token_probs}
\end{figure*}

We conducted five long-horizon tasks, each consisting of several skills commonly used in real-life scenarios, as detailed in Figure \ref{fig:room} and Table \ref{tab:scene}, to evaluate the performance of the proposed system in real-world settings.
Each task was accompanied by a natural language instruction $i$, and the robot initiated from the initial state at the beginning of each task.
The experiments were carried out using the Tiago++ \cite{pages2016tiago} mobile robot, which is equipped with wheels and two 7-degree-of-freedom arms. The robot features an integrated RGB-D camera on its head, which is utilized for perception throughout the tasks. 
For motion planning of the robotic arms, we employed Pinocchio \cite{carpentier2019pinocchio}. 
High-level planning and promptable predicate checking were performed using the gpt-4o \cite{openai2024gpt4technicalreport} model.
We used three computers: one for controlling the robot with the ROS Noetic system and others for generating navigation manipulation feedback. The robot employed the \verb|adjust| action to modify its position, head tilt, and pane orientation several times to make the manipulated object feasible to grasp if certain manipulation feasibility predicates were not met. 
Seven prompts are used for promptable predicates identification, PDDL state and goal generation, and task planning.
For other failed predicates, the robot could always use the \verb|alert| action to report issues to a human operator.



\subsection{Mobile Manipulation Tasks with the LLM Planner}

\noindent\textbf{Planning results.} We investigated the capability of the LLM planner to perform zero-shot planning for long-horizon tasks based on the robot's perception and a pre-built voxel map. We recorded the actions and constructed PDDL problems for five long-horizon tasks, with the results for the ``Take jacket'' task illustrated in Figure \ref{fig:plans-box}. Given the instruction, ``Pick the paper box on the wooden table and place it on the black table,'' and with the 2D and 3D images of the current state, we first construct the current state and goal state with predicates defined above.
Then, we filter and integrate all possible predicate-object pairs, which are utilized to extract the knowledge of the environment, by inputting the designed prompt, available predicates and objects, and user instructions into the LLM planner.
The promptable and grounding mechanism-based predicates are utilized to generate observations (e.g., the paper box is \verb|on| the wooden table), and feasibility feedback (e.g., \verb|find| a path to the paper box).
For instance, when the robot is \verb|at| the paper box (as determined by the visiting records) and the paper box is \verb|detected| (as determined by LangSAM), \verb|graspable| (as determined by filtering the grasping poses of the paper box generated by GraspNet), and \verb|reachable| (determined by the reachability map), while the robot is not \verb|holding| it, then the LLM planner will generate action primitive \verb|grasp| with its parameter \verb|paper_box| as the optimal action to execute. We list the actions and PDDL problems generated for the other four long-horizon tasks on the website. 

\noindent\textbf{Optimal actions.} To evaluate the optimality of the actions generated by the LLM planner during the task, we collected the token probabilities of four action candidates $\{ a_t^1, \cdots, a_t^4 \}$ generated for each action $a_t$, where $t \in \{1:H\}$, across five long-horizon tasks. The results are presented in Figure \ref{fig:token_probs}. The token probabilities exhibit variance due to the inherent uncertainty in the LLM's perception of the current state. 
Additionally, we observed that as the task progresses, the token probabilities of the four candidates converge, due to the reduction in task uncertainty as more information about the task and environment is provided to the LLM planner.
Furthermore, we observed that as the task progresses, the token probabilities of candidate actions become increasingly consistent and higher, corresponding to a decrease in task uncertainty as more information about the task and the environment is provided to the LLM planner. 

\begin{table}[tbp]
  \centering
  \begin{tabular*}{\linewidth}{@{\extracolsep{\fill}} cccccc }
    \toprule
    Tasks & \makecell{Place\\box} & \makecell{Take\\jacket} & \makecell{Take\\pill box} & \makecell{Insert\\book} & \makecell{Lift\\bucket} \\
    \midrule
    Navigation & 27 & 24 & 32 & 88 & 32 \\
    Manipulation & 179 & 75 & 112 & 208 & 256 \\
    Planning & 106 &	113 &	158 &	145 &	124 \\
    \bottomrule
  \end{tabular*}
  \caption{The temporal duration (measured in seconds) used for the completion of tasks in a single successful execution.}
  \label{tab:time}
\end{table}

\noindent\textbf{Temporal duration.} We also recorded the navigation, manipulation, and planning times used to complete these long-horizon tasks in the real world, with the results presented in Table \ref{tab:time}. 
These results demonstrate that our method can accomplish these tasks within a reasonable time.

\subsection{Ablation Study on Feasibility and Optimality}
\begin{figure}[t]
    \centering
    \includegraphics[width=\columnwidth]{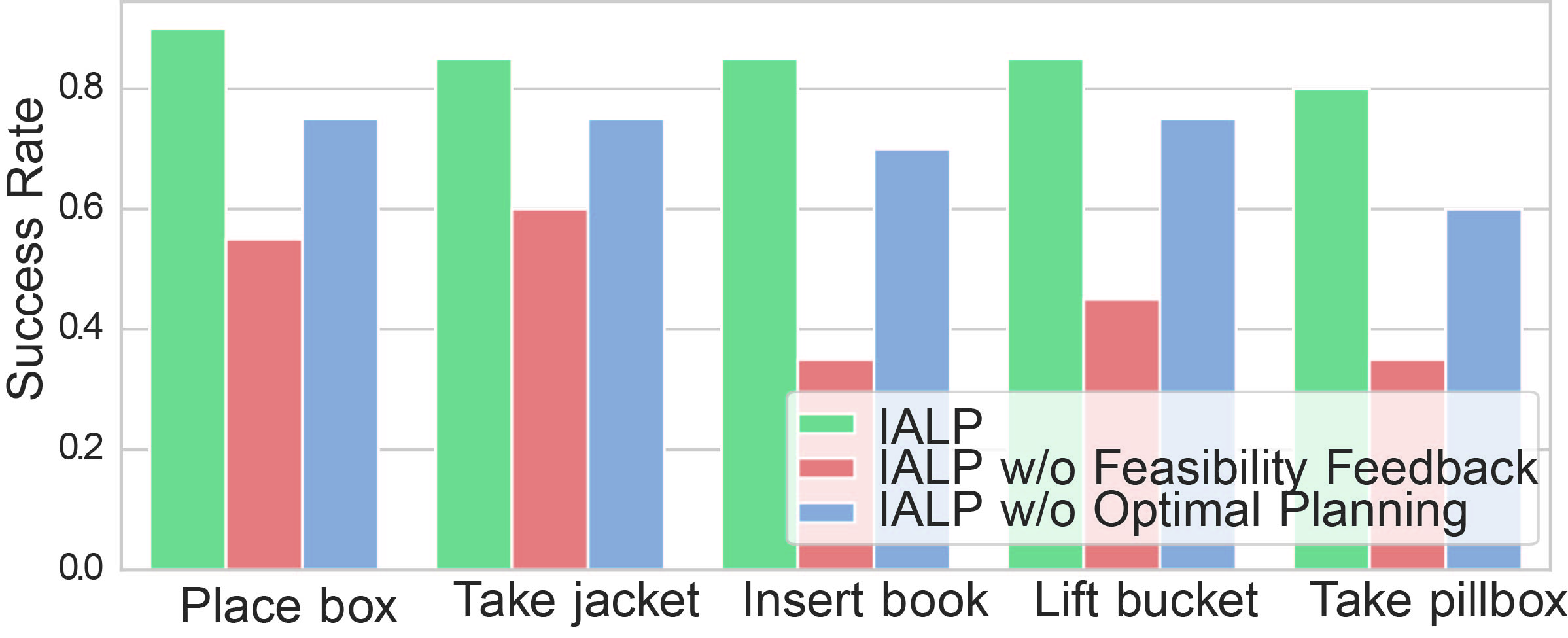}
    \caption{The success rate of IALP compared with that of IALP without feasibility feedback and without optimal selection, respectively.}
    \label{fig:success}
\end{figure}

To evaluate the impact of feasibility feedback and optimal selection on system performance, we conducted two ablation experiments, excluding either the feasibility feedback or optimal selection component separately to isolate their individual contributions to the overall system. All methods employed GPT-4o as the underlying planner.
We generated 20 sets of observations for each task in different settings. The predicates in each set of observations were randomly assigned, i.e., set to True with a probability of 50\%. We then recorded the success cases of the LLM planner generating executable actions, as shown in Figure \ref{fig:success}. The results indicate that IALP achieves a success rate of over 80\% in all long-term tasks. For the system without feasibility feedback (labeled as \textbf{IALP w/o Feasibility Feedback}), it encounters difficulty in generating feasible actions due to the removal of feasibility predicates such as \verb|find| and \verb|reachable| from the PDDL description. As a result, the success rate is substantially lower than that of other configurations.
For the system without optimal selection, denoted as \textbf{IALP w/o Optimal Selection}, a relatively high success rate is still maintained because feasibility checks are applied to all action generation, which ensures a certain degree of effectiveness despite the absence of optimal selection.

\subsection{Discussion}
\begin{figure}[t]
    \centering
    \includegraphics[width=1.0\columnwidth]{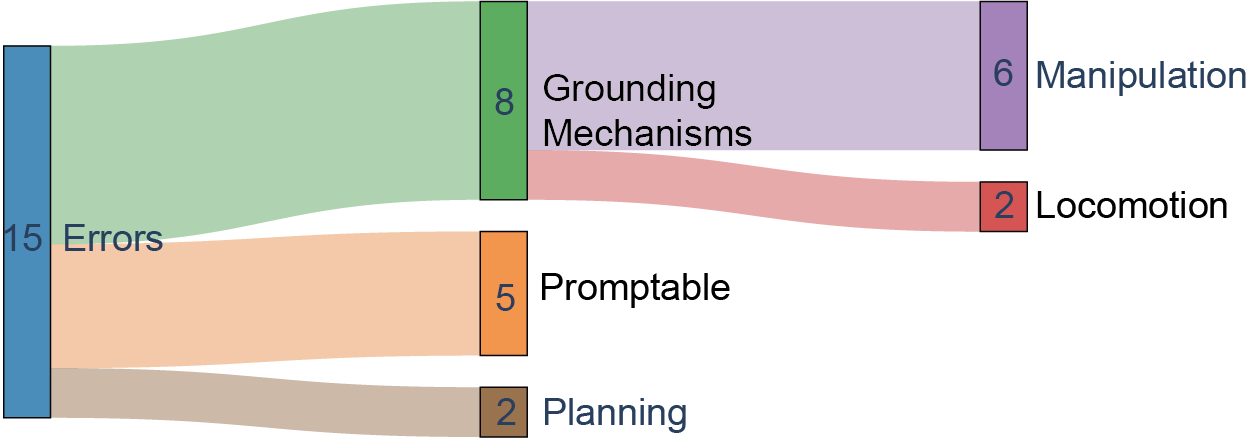}
    \caption{All failure cases of predicate checking in the real-world experiments across five long-horizon tasks.}
    \label{fig:errors}
\end{figure}

To investigate the types of failure cases in real-world experiments, we conducted 20 trials for each task within a real-world environment and recorded all occurring errors. All instances of predicate-checking failures were systematically aggregated and classified into three categories: planning, promptable, and grounding mechanisms failures. The results are presented in Figure \ref{fig:errors}. Planning failures occur when the planner fails to generate the correct action sequence. In contrast, promptable and grounding mechanism failures arise when predicates listed in Table \ref{tab:predicates} yield incorrect outputs, such as incorrect grasp poses for graspable objects. While 15 errors out of 100 may appear insignificant, they represent a considerable workload in real-world hardware experiments compared with numerical simulations due to factors such as hardware issues, noise, and physical limitations.

\section{Conclusion}
We propose IALP, a framework that leverages promptable and grounding mechanism-based predicates to construct an informative PDDL problem to represent task-relevant information of the current state. This enables the LLM planner to generate feasible and optimal actions in a closed-loop manner.
Despite achieving the anticipated performance, our system has certain limitations. IALP is dependent on a pre-defined primitive action library, which restricts the range of task instructions and impedes the generalization of existing actions.
Future work will focus on developing a more generalized action model capable of learning multiple skills and enhancing the dynamic reasoning ability of VLMs.

\section*{Acknowledgments}
This work is supported in part by the Research Grants Council of Hong Kong under grant number C4042-23GF, and in part by the PolyU-EIT Collaborative PhD Training Programme under application number 220724983.

\bibliography{main}

\onecolumn
\appendix

This appendix provides further elaboration on the implementation of IALP. Section \ref{sec:derivation} presents the detailed derivations of our objective, feasibility score, and optimality score. Section \ref{sec:exp} offers comprehensive experimental details, including the design of actions, predicates, and the domain as described by PDDL. Section \ref{sec:imp} lists all the prompts utilized in our experiments, consisting of those for checking promptable predicates, generating PDDL problems, and serving as the LLM planner.

\section{Derivations} \label{sec:derivation}
We provide the derivatios of our objective defined in \textbf{Problem Formulation} section. The objective at time step $t \in \{1:H\}$ can be expressed as the joint probability of skill sequence $a_{t:H}$ and binary rewards $r_{t:H}$ given the instruction $i$ and the current state $s_t$:

\begin{align}
\label{eq:pf-2}
  p(a_{t:H}, r_{t:H}|i, s_t) &= p(a_{t:H} \mid i, s_t) p(r_{t:H} \mid i, s_t, a_{t:H}) \tag{conditional probability} \\
                           &= \prod_{x=t}^H p(a_x \mid i, s_t, a_{t:x-1}) \prod_{x=t}^H p(r_x \mid i, s_t, a_{t:x}, r_{t:x-1}) \tag{conditional probability} \\
\end{align}

\subsection{Optimality}
\label{sec:op}
Here, we extract the probability of action $a_x, x \in \{t:H\}$, generated at time step $t$ from the first term of Equation \ref{eq:pf-2}.

\begin{align}
  p(a_x \mid i, s_t, a_{t:x-1}) &= \int p(a_x \mid i, s_{t:x}, a_{t:x-1}) p(s_{t+1:x} \mid i, s_t, a_{t:x-1}) ds_{t+1:x} \label{eq:fb-1} \\
                                           &= \mathbb{E}_{s_{t+1:x}} [p(a_x \mid i, s_{t:x}, a_{t:x-1})] \label{eq:fb-2}\\
                                           &\approx p(a_x \mid i, s_{t:x}, a_{t:x-1}) \label{eq:fb-4}
\end{align}

The Equation \ref{eq:fb-4} is the feasibility score. By computing a single sample of Monte-Carlo estimation under the future state trajectory $s_{t+1:x}$, where $s_y = \mathcal{T}(s_{y-1}, a_{y-1}), y \in \{t+1, x\}$, we ensure that the future states $s_{t+1:x}$ correspond to a successful execution of the running plan $a_{t:x-1}$, i.e., achieving positive rewards $r_{t:x-1}$.
The key insight is that future states $s_{t+1:x}$ are only ever sampled after performing feasibility planning to maximize the success probability of the running plan $a_{t:x-1}$.
Specifically, the probability of action $a_t$ generated at time step $t$ is approximate as $p(a_t \mid i, s_t)$, that is, $a_t$ only depends on current state $s_t$ and the user instruction $i$.

\subsection{Feasibility}
\label{sec:fb}
Here, we extract the probability of reward $r_x, x \in \{t:H\}$, generated at time step $t$ from the second term of Equation \ref{eq:pf-2}.

\begin{align}
  p(r_x \mid i, s_t, a_{t:x}, r_{t:x-1}) &= \int p(r_x \mid i, s_{t:x}, a_{t:x}, r_{t:x-1}) p(s_{t+1:x} \mid i, s_t, a_{t:x}, r_{t:x-1}) ds_{t+1:x} \label{eq:op-1} \\
                                           &= \mathbb{E}_{s_{t+1:x}} [p(r_x \mid i, s_{t:x}, a_{t:x}, r_{t:x-1})] \label{eq:op-2}\\
                                           &\approx \mathbb{E}_{s_{t+1:x}} [p(r_x \mid i, s_{t:x}, a_{t:x})] \label{eq:op-3}\\
                                           &\approx p(r_x \mid i, s_{t:x}, a_{t:x}) \label{eq:op-4}
\end{align}

The reward $r_x$ generated at time step $t$ is independent of the instruction $i$. As described in Appendix \ref{sec:op}, we can make an independence assumption on previous rewards $r_{t:x-1}$ between Equation \ref{eq:op-2} and Equation \ref{eq:op-3}.
Specifically, the probability of reward $r_t$ generated at time step $t$ is approximate as $p(r_t \mid i, s_t, a_t)$, that is, $r_t$ only depends on the user instruction, current state $s_t$ and action $a_t$.

\newpage
\section{Experiment Details} \label{sec:exp}

\subsection{PDDL actions}

The following are the primitive actions used in our experiments, as described by PDDL. It defines the primitive function $\psi$, the parameters $\alpha$, the precondition predicates $\alpha^{p}$ and the effect predicates $\alpha^{e}$.
Additionally, three supplementary actions are introduced: \verb|adjust|, which is used to adjust the robot (head) if the object is not reachable, \verb|alert|, which is employed to notify the human operator of an error that the robot cannot resolve given the current knowledge, and \verb|stop|, which is used to halt the robot once the task is completed.

\begin{lstlisting}
(:action move ; move the robot to the object
    :parameters (?obj - locatable)
    :precondition (and
        (find ?obj)
    )
    :effect (and
        (at ?obj)
    )
)
(:action scan ; scan the object
    :parameters (?obj - locatable)
    :precondition (and
        (at ?obj)
        (not (detected ?obj))
    )
    :effect (and
        (detected ?obj)
    )
)
(:action grasp ; grasp the object
    :parameters (?obj - locatable)
    :precondition (and
        (at ?obj)
        (detected ?obj)
        (graspable ?obj)
        (reachable ?obj)
    )
    :effect (and
        (holding ?obj)
    )
)
(:action place ; place the object on the place
    :parameters (?obj - locatable ?pla - locatable)
    :precondition (and
        (at ?place)
        (holding ?obj)
        (placeable ?pla)
    )
    :effect (and
        (on ?obj ?pla)
    )
)
(:action pull ; pull the object
    :parameters (?obj - locatable)
    :precondition (and
        (at ?obj)
        (graspable ?obj)
        (not (opened ?obj))
    )
    :effect (and
        (opened ?obj)
    )
)
(:action push ; push the object
    :parameters (?obj - locatable)
    :precondition (and
        (at ?obj)
        (graspable ?obj)
        (opened ?obj)
    )
    :effect (and
        (not (opened ?obj))
    )
)
(:action lift ; lift the object
    :parameters (?obj - locatable)
    :precondition (and
        (at ?obj)
        (holding ?obj)
    )
    :effect (and
    )
)
(:action rotate ; rotate the object
    :parameters (?obj - locatable)
    :precondition (and
        (at ?obj)
        (holding ?obj)
    )
    :effect (and
    )
)
(:action reach ; reach the pose
    :parameters (?pose - locatable)
    :precondition (and
        (reachable ?pose)
    )
    :effect (and
    )
)
(:action adjust ; adjust the robot if the object is not reachable, graspable, or placeable
    :parameters (?obj - locatable)
    :precondition (or
        (not (reachable ?obj))
        (not (graspable ?obj))
        (not (placeable ?obj))
    )
    :effect (and
    )
)
(:action alert ; alert when there's an error the robot cannot solve based on the current state
    :parameters ()
    :precondition (and 
    )
    :effect (and
    )
)
(:action stop ; stop the robot when the taks is done
    :parameters ()
    :precondition (and
    )
    :effect (and
    )
)
\end{lstlisting}

\subsection{PDDL domain}
We introduce the room domain described by PDDL in this section. It contains two major parts: predicates and actions.
\begin{lstlisting}
(define (domain room)
    (:requirements :strips :fluents :durative-actions :timed-initial-literals :typing :conditional-effects :negative-preconditions :duration-inequalities :equality :disjunctive-preconditions)
    (:types
        locatable - object
        floor - object
        bot table coat_stand cloth box - locatable
        robot - locatable
        pose - locatable
    )
    (:predicates
        (:predicates ;todo: define predicates here
            (on ?obj1 - locatable ?obj2 - locatable) ; if the object is on another object
            (in ?obj - locatable ?obj2 - locatable); if the object1 is inside others
            (holding ?obj - locatable) ; if the arm is holding the object
            (opened ?obj) ; if the object is opened
            (at ?obj - locatable) ; if the robot is at the object (near)
            (find ?obj - locatable); if has a path from the current state to the object
            (detected ?obj - locatable); if the object is detected
            (graspable ?obj - locatable); if the object is graspable
            (reachable ?obj - locatable); if the object is reachable
            (placeable ?obj - locatable); if the object is placeable
        )
    )
    (:functions
    )
    ; Actions
    ... Refer to Appendix B.1 ...
)
\end{lstlisting}

\subsection{Other mobile manipulation tasks results}
\begin{figure}[ht]
    \centering
    \includegraphics[width=\columnwidth]{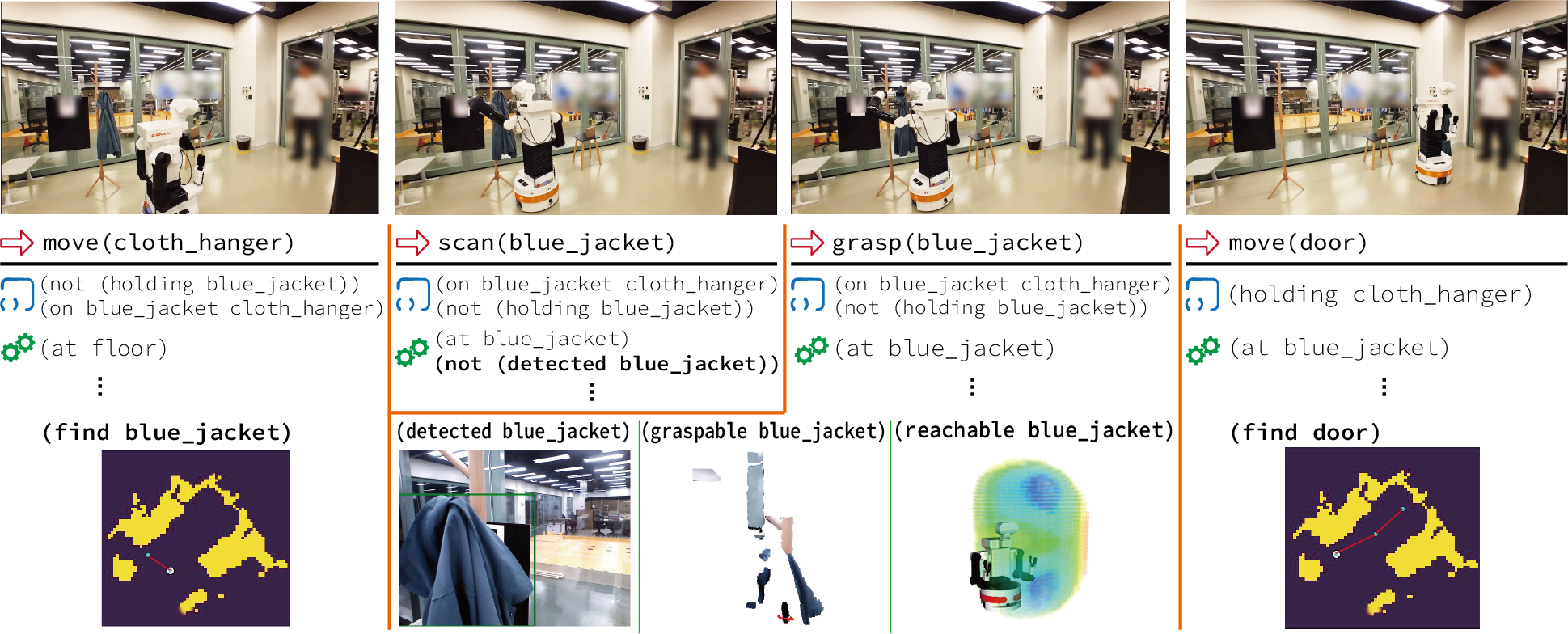}
    \caption{The states, feasibility feedback, and actions generated during the execution of the task Take jacket.}
    \label{fig:take_cloth}
\end{figure}

\begin{figure}[ht]
    \centering
    \includegraphics[width=\columnwidth]{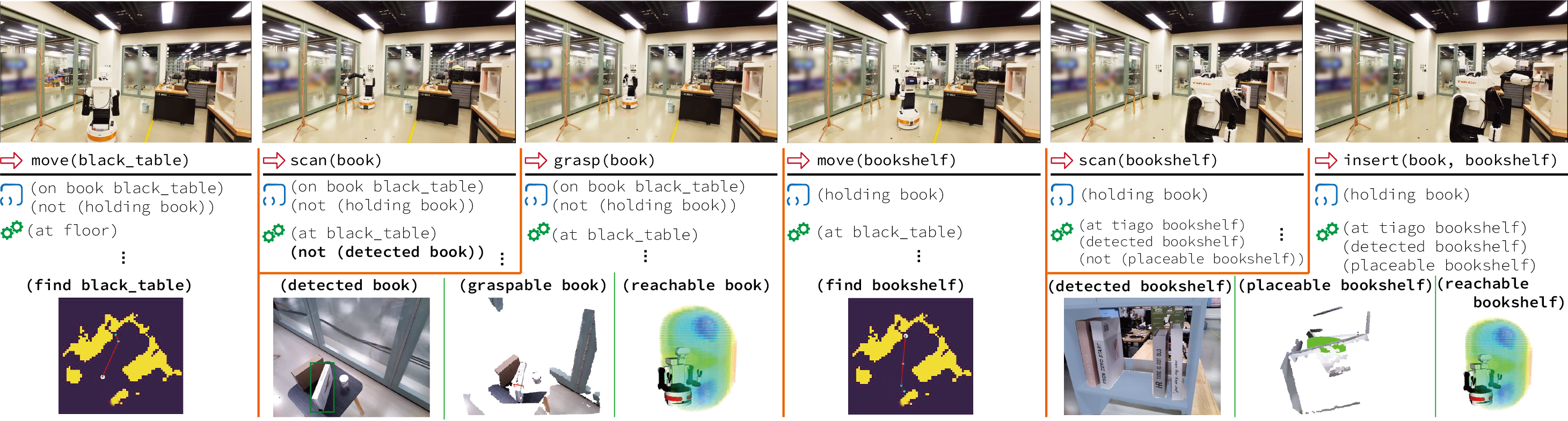}
    \caption{The states, feasibility feedback, and actions generated during the execution of the task Insert book.}
    \label{fig:take_book}
\end{figure}

\begin{figure}[ht]
    \centering
    \includegraphics[width=\columnwidth]{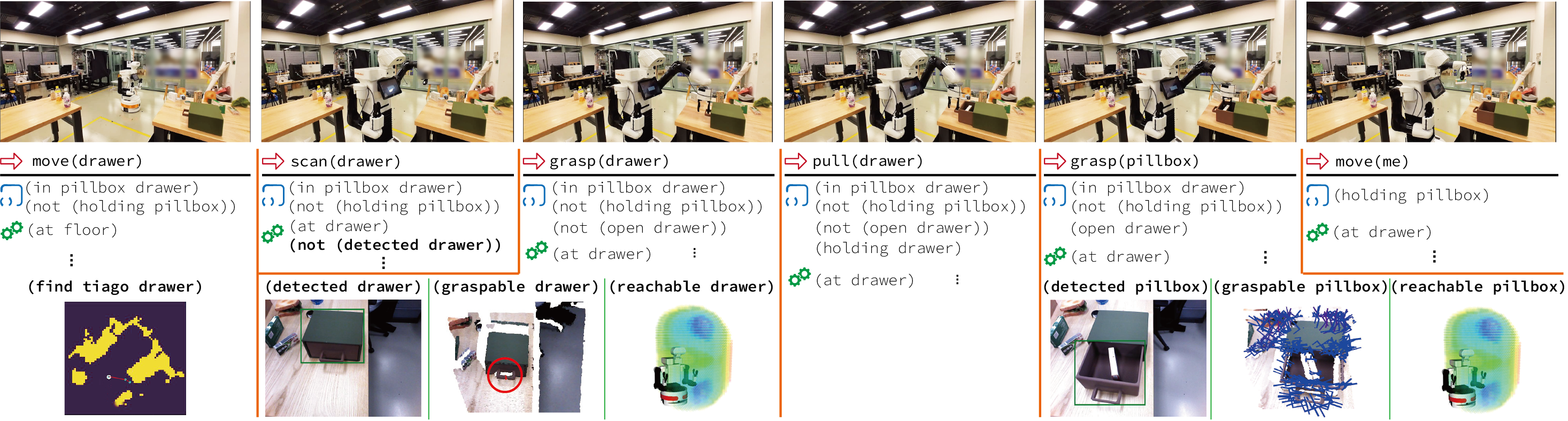}
    \caption{The states, feasibility feedback, and actions generated during the execution of the task Take pillbox.}
    \label{fig:take_pillbox}
\end{figure}

\begin{figure}[ht]
    \centering
    \includegraphics[width=\columnwidth]{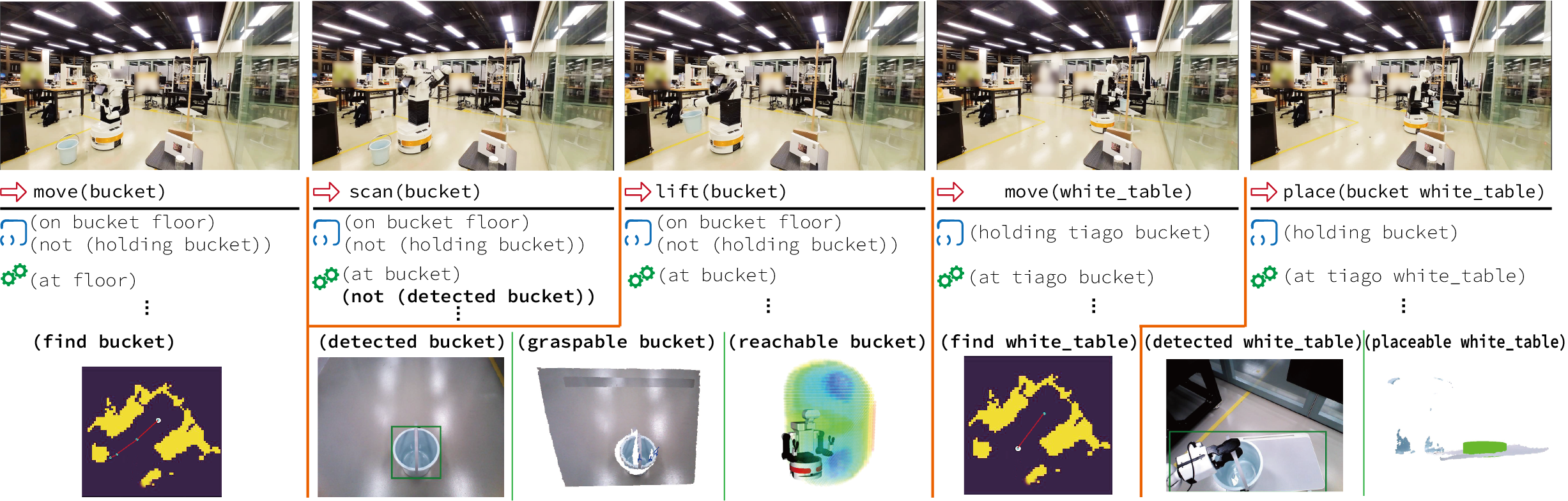}
    \caption{The states, feasibility feedback, and actions generated during the execution of the task Lift bucket.}
    \label{fig:lift_bucket}
\end{figure}

We list the results of the other four mobile manipulation tasks: Take jacket, Insert book, Take pillbox, and Lift bucket, in Figure \ref{fig:take_cloth}, \ref{fig:take_book}, \ref{fig:take_pillbox} and \ref{fig:lift_bucket}, respectively. We refer to Supplementary files for code and data generated during experiments.


\newpage
\section{Implementation Details} \label{sec:imp}
In this section, we list all the prompts used in our experiments. Specifically, in Section \ref{sec:pb}, we present the prompts used to check promptable predicates. In Section \ref{sec:pob}, we provide the prompts employed to extract task-related objects and determine which predicates will be used to describe the state of the task. In Section \ref{sec:pgo}, we outline the prompts used to construct the goal description. Finally, in Section \ref{sec:taskP}, we detail the prompts used to generate actions to complete the long-horizon task.

\subsection{Promptable predicates} \label{sec:pb}
\noindent\textbf{on? or in?} Checking the spatial relationship between two objects.
\begin{lstlisting}
Please determine the spatial relationship between [OBJECT1] and [OBJECT2] based on the given instruction. Return "on" or "in" only.
The instruction: [INSTRUCTION]
\end{lstlisting}

\noindent\textbf{open?} Checking if the object, like the drawer, is opened.
\begin{lstlisting}
Please check if [OBJECT] is opened in the image. Return True or False only. If there is no [OBJECT] in the image, just return False.
\end{lstlisting}

\noindent\textbf{holding?} Checking if the robot is holding the object.
\begin{lstlisting}
Please determine if the robotic arm is holding the [OBJECT] in the image. Return True or False only. If there is no [OBJECT] in the image, just return False.
\end{lstlisting}

\subsection{Prompts to generate observation} \label{sec:pob}

\noindent\textbf{Extract objects} Extracting task-related objects in current task.
\begin{lstlisting}
Please extract the object name and related object name from the given instruction. The related object name will help humans to find the object.  Also, extract other object names that may be related to finishing the instruction, such as the target position-related objects. You need to concatenate multi-word object names by using "_." For example, the "black table" in the instruction should be converted to "black_table." The answer should be in JSON format without markdown code block triple backticks:

{
    "object_name": "str",
    "related_object_name": "str",
    "other_object_names": [
        "str",
        "str"
    ]
}

The instruction: [INSTRUCTION]
\end{lstlisting}

\noindent\textbf{Decide which predicates will be used} Listing all possible predicates that necessary to check the current state.
\begin{lstlisting}
You are required to detect the observation of the current environment by using the predicates and related objects provided below. You need to give all predicates and objects necessary to check the current state so that the robot can choose the best action. You don't need to verify each predicate. Every predicate can be used multiple times. Do NOT return markdown code block triple backticks.

Instruction: [INSTRUCTION]
The possible ?obj could be: [OBJECTS]

The predicate candidates are:
"""
predicates ...
"""

The possible actions are:
"""
actions ...
"""

The output is formatted as:
"""
(on box table)
(holding robot box)
...
"""
\end{lstlisting}

\subsection{Prompts to generate goals} \label{sec:pgo}
\begin{lstlisting}
Based on the instruction, you are using the following predicates to generate the goal of the PDDL problem. The robot's name is Tiago.

Instruction: [INSTRUCTION]

The predicate candidates are:

"""
Refer to Appendix B.1 ...
"""

You can use (and ) and (or ) to combine the goal predicates. Please only return answers without any explanation. Do not return markdown code wrappers.

Here's an example:
[user]
Instruction: Pick up the box on the table and place it on the black table.
[assistant]
(on box black_table)
Example finished.

Here's what I give to you:
Instruction: [INSTRUCTION]
\end{lstlisting}

\subsection{Task planning with the LLM planner} \label{sec:taskP}
\noindent\textbf{System role}
\begin{lstlisting}
[user]
You are an excellent interpreter of human instructions for daily tasks. Given an instruction and information about the working environment, you break it down into a sequence of robotic actions. Please do not begin working until I say "Start working." Instead, simply output the message "Waiting for next input." Understood?

[assistant]
Understood. Waiting for the next input.
\end{lstlisting}

\noindent\textbf{Environments}
\begin{lstlisting}
[user]
Information about environments, objects, and tasks is given as a PDDL function.
The Planning Domain Definition Language (PDDL) is a domain-specific language designed for the Benchmark for creating a standard for Artificial Intelligence (AI) planning.

Here's the domain description you used:
"""
PDDL domain, refer to Appendix B.1
"""
The =:action= blocks define all the action/subtasks used for completing the task.

Later, you will receive the task/problem defined by PDDL and the above domain.
Here's an example:
"""
Example of PDDL problem, refer to Appendix B.2
"""

The =:init= block defines the current observation of the environment.
The =:goal= block defines the goal of the task.

You need to take action from the current state, not from the start. If the current task is over and no action is needed for the robot, you can use the "stop" action. If the robot doesn't know what to execute in its current state, for example, it cannot find the target object, you can use the "alert" action and stop the robot. If the object is not reachable, graspable, or placeable after the scan, you can use the "adjust" action to adjust the robot's pose to make it reachable, graspable, or placeable.

-------------------------------------------------------
The texts above are part of the overall instruction. Do not start working yet:
[assistant]
Understood. Waiting for the next input.
\end{lstlisting}

\noindent\textbf{Observation input}
\begin{lstlisting}
Start working. Resume from the environment below.

The instruction is as follows:
"""
[INSTRUCTION]
"""

The action executed last time is as follows:
"""
[ACTION]
"""

The observation of the current environment is as follows:
"""
[OBSERVATION]
"""
\end{lstlisting}

\end{document}